\documentclass{article}

\usepackage[final]{neurips_2019}

\usepackage[utf8]{inputenc} 
\usepackage[T1]{fontenc}    
\usepackage{hyperref}       
\usepackage{url}            
\usepackage{booktabs}       
\usepackage{amsfonts}       
\usepackage{nicefrac}       
\usepackage{microtype}      
\usepackage{amsmath}
\usepackage{graphicx}
\usepackage{subfig}
\usepackage{float}

\title{Reconsidering Analytical Variational Bounds for Output Layers of Deep Networks}

\author{%
Otmane Sakhi\\
  Criteo AI Lab, Paris\\
\And
Stephen Bonner \\
Department of Computer Science\\
Durham University, Durham
\And
David Rohde\\
Criteo AI Lab, Paris\\
\And
Flavian Vasile \\
Criteo AI Lab, Paris\\
}

\newcommand{\bX}{\boldsymbol{X}}
\newcommand{\by}{\boldsymbol{y}}

\newcommand{\bmu}{\boldsymbol{\mu}}
\newcommand{\bSigma}{\boldsymbol{\Sigma}}

\newcommand{\bPsi}{\boldsymbol{\Psi}}

\newcommand{\bomega}{\boldsymbol{\omega}}

\newcommand{\bI}{\boldsymbol{I}}
\newcommand{\bzero}{\boldsymbol{0}}
\newcommand{\brho}{\boldsymbol{\rho}}

\newcommand{\bv}{\boldsymbol{v}}
\newcommand{\bbeta}{\boldsymbol{\beta}}

\begin{document}

\maketitle

\begin{abstract}
    The combination of the re-parameterization trick with the use of variational auto-encoders has caused a sensation in Bayesian deep learning, allowing the training of realistic generative models of images and has considerably increased our ability to use scalable latent variable models. The re-parameterization trick is necessary for models in which no analytical variational bound is available and allows noisy gradients to be computed for arbitrary models. However, for certain standard output layers of a neural network, analytical bounds are available and the variational auto-encoder may be used both without the re-parameterization trick or the need for any Monte Carlo approximation. In this work, we show that using Jaakola and Jordan bound, we can produce a binary classification layer that allows a Bayesian output layer to be trained,  using the standard stochastic gradient descent algorithm. We further demonstrate that a latent variable model utilizing the Bouchard bound for multi-class classification allows for fast training of a fully probabilistic latent factor model, even when the number of classes is very large.
\end{abstract}

\section{Introduction}

Imagine we have $N$ conditionally independent draws of $\bX$ from a model and we would like to consider a variational approximation under the variational posterior $q(\cdot|\eta)$. In a number of interesting cases a variational bound is not available directly, but a bound becomes available if we augment the model with $N$ additional variational parameters $\zeta_1, ..., \zeta_N$.  Specially we are interested in cases where the lower bound has the following form:

\[
\log p(\bX_1, ..., \bX_N) \ge \mathcal{H}(\eta) + \sum_n^N \mathcal{F}(\bX_n,\zeta_n, \eta) = \mathcal{L}
\]

Typically the sum over the data $\mathcal{F}(\bX_n,\zeta_n, \eta)$ is a bound on the likelihood, and $\mathcal{H}(\eta)$ is the negative Kullback Leibler divergence.
This bound contains variational parameters $\zeta_1,...,\zeta_N$ that must be optimized, but that increase in dimension proportional to the number of records. This required increase in dimension can make large data inference intractable, instead we consider using a variational auto-encoder $\zeta_n = f_\Xi(\bX_n)$ which reduces the dimension of the optimization problem to $(\eta,\Xi)$. Alternatively, if a variational EM algorithm exists, then there may be a tractable known expression for $\zeta_n = f(\bX_n)$ which does not require learning the parameters of an auto-encoder, simplifying the learning process.

Due to the sum structure, it is also possible to use Stochastic Gradient Descent (SGD) or the Robbins Monro algorithm by considering a noisy version of the bound. Combining these two steps results in the following noisy (but fast) objective:

\[
\hat{\mathcal{L}}(x_n,\eta,\Xi) = \frac{1}{N} \mathcal{H}(\eta) + \mathcal{F}(\bX_n, f_\Xi(\bX_n), \eta)
\]

While this method uses a variational auto-encoder $f_\Xi(\cdot)$ (if no EM step is available) unlike \cite{kingma2013auto,kingma2015variational} it uses an analytical lower bound in place of the re-parameterization trick. Analytical bounds are available for the output layers of many deep neural networks, including binary classifiers, categorical classifiers, Poisson count models and they do not require Monte Carlo samples to be drawn from the variational distribution.

The key advantage to this method is that training may be done by simply replacing the output layer with the auto-encoding analytical bound. This can then be training using standard SGD without the re-parameterization trick and with less noise in the gradients. Our method does however employ additional analytical bounds that the Kingma and Welling algorithm does not, meaning an additional approximation is used.

Models that have this form might be fully Bayesian treatments of latent variable models that are immediately recognizable as models that can be treated with variational auto-encoders, but there are also models, including Bayesian logistic regression, which also can be put into this form; the use of a variational auto-encoder in this setting doesn't have the usual ``auto-encoding'' interpretation of reconstructing the input data.

The methodology also can be applied to solve integrated maximum likelihood problems for latent variable models of the following form:

\[
p(\bX_n|\theta) = \int p(\bX_n,z_n|\theta) dz_n \ge e^{\mathcal{F}(\zeta_n,\theta)}.
\]

Using a variational auto-encoder $\zeta_n = f_\Xi(\bX_n)$, we bound the dimension to the size of $\theta,\Xi$ and can obtain noisy estimates of the bound:

\[
\mathcal{\hat{L}}(\bX_n,\theta,\Xi) =   \mathcal{F}(f_\Xi(\bX_n),\theta)
\]

\section{Binary Classification Output Layer}

We consider the logistic regression model:

\[
\bbeta \sim \mathcal{N}(\bmu_{\beta},\bSigma_{\beta}), \hspace{2cm} y_n|\bX_n,\bbeta \sim {\rm Bernoulli}(\sigma(\bX_n^T \bbeta)).
\]

We can also view $\bX$ as the outputs from the second last layer of a deep network, and $\beta$ as the weights of the final layer. We can bound the the posterior using both the ELBO and the Jaakola and Jordan bound \cite{jaakkola1997variational} \cite{ormerod2010explaining}, and with respect to a variational distribution $q(\cdot)$ which we make a normal distribution of the form $\bbeta \sim \mathcal{N}(\bmu_q,\bSigma_q)$:

\begin{align*}
    \rm ELBO & =   - {\rm KL}(\bmu_q,\bSigma_q,\bmu_{\beta},\bSigma_\beta) + \mathop{\mathbb{E}_q[\log p(D|\beta)]}\\
  & = - {\rm KL}(\bmu_q,\bSigma_q,\bmu_{\beta},\bSigma_\beta) + \sum_n^N  y_n \bX_n^T \bmu_q - \mathop{\mathbb{E}_q[\log(1 + \exp(\bX_n^T \bbeta))]}\\
  & \ge - {\rm KL}(\bmu_q,\bSigma_q,\bmu_{\beta},\bSigma_\beta) + \sum_n^N  y_n \bX_n^T \bmu_q - \frac{1}{2} \bX_n^T \bmu_q + \max_{\zeta_n} ( A(\zeta_n) ( (\bX_n^T \bmu_q)^2 + \bX_n^T \bSigma_q X_n ) + C(\zeta_n))
  \end{align*}
  
\newpage

where:
\[
A(\zeta) = -\tanh(\zeta/2)/(4 \zeta) \hspace{2cm} C(\zeta) = \zeta/2 -\log(1+e^\zeta) +  \zeta \tanh(\zeta/2) /4. 
\]
and 
\[
{\rm KL}(\bmu_q,\bSigma_q,\bmu_{\beta},\bSigma_\beta) =  { 1 \over 2 }\log \frac{|   \bSigma_{\beta} |}{| \bSigma_q |} + { 1 \over 2 }  \operatorname{tr} \left( \bSigma_{\beta}^{-1} \bSigma_q \right) + { 1 \over 2 } \left( \bmu_{\beta} - \bmu_q\right)^{\rm T} \bSigma_{\beta}^{-1} ( \bmu_{\beta} - \bmu_q ) + { k \over 2 } 
\]

We can simplify the problem by finding a function $\zeta_n = f_\Xi(\bX_n,\bmu_q,\bSigma_q)$, while we may learn such an ``auto-encoding'' function with a deep net, in this case we can simply substitute the appropriate step from the variational EM algorithm giving:

\[
\zeta_n  = f(\bX_n)  = \sqrt{\bX_n^T \bSigma_q \bX_n + (\bX_n^T \bmu_q)^2}
\]

\begin{align*}
& \mathcal{L}(\bX,\by,\bmu_q,\bSigma_q)  =  - {\rm KL}(\bmu_q,\bSigma_q,\bmu_{\beta},\bSigma_\beta)  \\
  &  + \sum_n^N  y_n \bX_n^T \bmu_q + A( f(\bX_n) ) ( ( \bX_n^T \bmu_q)^2 + \bX_n^T \bSigma_q \bX_n ) - \frac{1}{2} \bX_n^T \bmu_q + C( f(\bX_n) ),\\    
\end{align*}

Our likelihood lower bound becomes deterministic and can be easily optimised by any SGD based method.




\section{Multiclass Latent Variable Model}

Consider the following latent variable model, which models the behavior of $U$ user sessions interacting with $P$ products, where the number of events for the session of user $u$ is denoted $T_u$.  The purpose of the model is to identify products that are of similar type that are often viewed together in the same session.  The model over a single session has the following form:
\[
\bomega_u \sim \mathcal{N}(\bzero,\bI), \hspace{2cm} v_{u,1},...,v_{u,T_u} \sim {\rm categorical}({\rm softmax}(\bPsi \bomega_u + \brho)).
\]

The log probability can be written:

\begin{align*}
    \log ~  & p(\bv,\bomega|\bPsi,\brho)  = \sum_u^U \left(  \sum_t^{T_u} \bPsi_{v_{u,t}}
              \bomega_u + \brho_{v_{u,t}} \right) \\
  &  -T_u \log\{ \sum_p^P \exp( \bPsi_p \bomega_u  +\brho_p)\}  - \frac{K}{2} \log( 2 \pi )-
                                         \frac{1}{2}  \bomega_{u}^T
                                         \bomega_{u},
  \end{align*}

we can bound the integrated log likelihood using the Bouchard bound \cite{bouchard2007efficient} \cite{rohde2019latent} with respect to a variational distribution $q(\cdot)$, which we parametrerize as a normal distribution such that $\bomega \sim \mathcal{N}(\bmu_q,\bSigma_q)$:

\begin{align*}
    &  \mathcal{L} = \sum_u  - \frac{K}{2} \log( 2 \pi )-  \frac{1}{2}  \{ \bmu_{q_u}^T \bmu_{q_u} + {\rm trace} (\bSigma_{q_u})  \} + \frac{1}{2} \log |2 \pi e \bSigma_{q_u} |\\
    & +    \sum_u^U \left(  \sum_t^{T_u} \bPsi_{v_{u,t}} \bmu_{q_u}  + \brho_{v_{u,t}} \right)  -T_u [ 
      a_u + \sum_p^P \frac{\bPsi_p \bmu_{q_u} +\brho_p - a_u - \xi_{u,p}}{2} \\
      & + \lambda_{\rm JJ}(\xi_{u,p}) 
        \{(\bPsi_p \bmu_{q_u}+\brho_p-a_u)^2 + \bPsi_p \bSigma_{q_u} \bPsi_p^T  - \xi_{u,p}^2  \} + \log(1 + e^{\xi_{u,p} })
      ],\\  
\end{align*}

where
\[
    \lambda_{\rm JJ}(\xi) = \frac{1}{2\xi} \left( \frac{1}{1+e^{-\xi}} - \frac{1}{2} \right).
\]

We then use the following variational auto-encoders:

\[
\bmu_{q_u} = g_\Xi^\mu (\bv_u), \hspace{1cm} \bSigma_{q_u} = g_\Xi^\Sigma (\bv_u), \hspace{1cm} a_u = g_\Xi^a (\bv_u).
\]

For $\xi_{u,p}$, rather than using the auto-encoder, we can use an explicit update (derived from the variational EM algorithm \cite{rohde2019latent}):
\[
\xi_{u,p} = g^\xi(\bv_u,p) = \sqrt{\bPsi_p \bSigma_{q_u} \bPsi_p^T + (\bPsi_p \bmu_{q_u} + \brho_p-a_u)^2 }
\]

Substituting the auto-encoders and update into the lower bound causes the optimization problem to be written as a finite sum over each of the $U$ time-lines, thus allowing SGD to be applied. More remarkably it also causes the denominator of the softmax to decompose into a sum over UP terms. This allows not only a fast computation of the bound by sampling individual records, but an even faster (and noiser) bound to be computed by also sampling a small subset of the P items involved in the partition function.  This can accelerate learning when P is large, which otherwise requires heuristics such as the famous but non-probabilistic word2vec (skipgram with negative sampling) algorithm \cite{word2vec}. Our proposed method is similar to \cite{raman2016ds} but our use of an auto-encoder means that plain SGD is all that is required.

The noisy lower bound becomes:
\begin{align*}
  &  \hat{\mathcal{L}}(v_{u,1},...,v_{u,T_u} ,s_1,...s_S,\Xi,\bPsi) \\
 & =   - \frac{K}{2 U} \log( 2 \pi )-  \frac{1}{2 U}  \{ g_\Xi^\mu (\bv_u)^T g_\Xi^\mu (\bv_u) + {\rm trace} (g_\Xi^\Sigma (\bv_u) )  \} + \frac{1}{2 U}  \log |2 \pi e g_\Xi^\Sigma (\bv_u) |\\
  & +     \left(  \sum_t^{T_u} \bPsi_{v_{u,t}} g_\Xi^\mu (\bv_u)  + \brho_{v_{u,t}} \right)  - T_u [ 
    g_\Xi^a (\bv_u) + \frac{P}{S} \sum_s^S \frac{\bPsi_{p_s} g_\Xi^\mu (\bv_u) +\brho_{p_s} - g_\Xi^a (\bv_u) - g^\xi(\bv_u,p) }{2} \\
    & + \lambda_{\rm JJ}(g^\xi(\bv_u,p) ) 
      \{(\bPsi_{p_s} g_\Xi^\mu (\bv_u) +\brho_{p_s}-g_\Xi^a (\bv_u) )^2 + \bPsi_{p_s} g_\Xi^\Sigma (\bv_u) \bPsi_{p_s}^T  - g^\xi(\bv_u,p)^2  \} \\
    &  + \log(1 + e^{ g^\xi(\bv_u,p) })],\\  
\end{align*}

where $v_{u,1},...,v_{u,T_u}$ are the items associated with session $u$ and $s_1,...,s_S$ are $S<P$ negative items randomly sampled.

\section{Experiments}

\subsection{Jaakola and Jordan Logistic Regression SGD}

\begin{figure}[H]
  \centering
  \subfloat{
    \includegraphics[width=60mm]{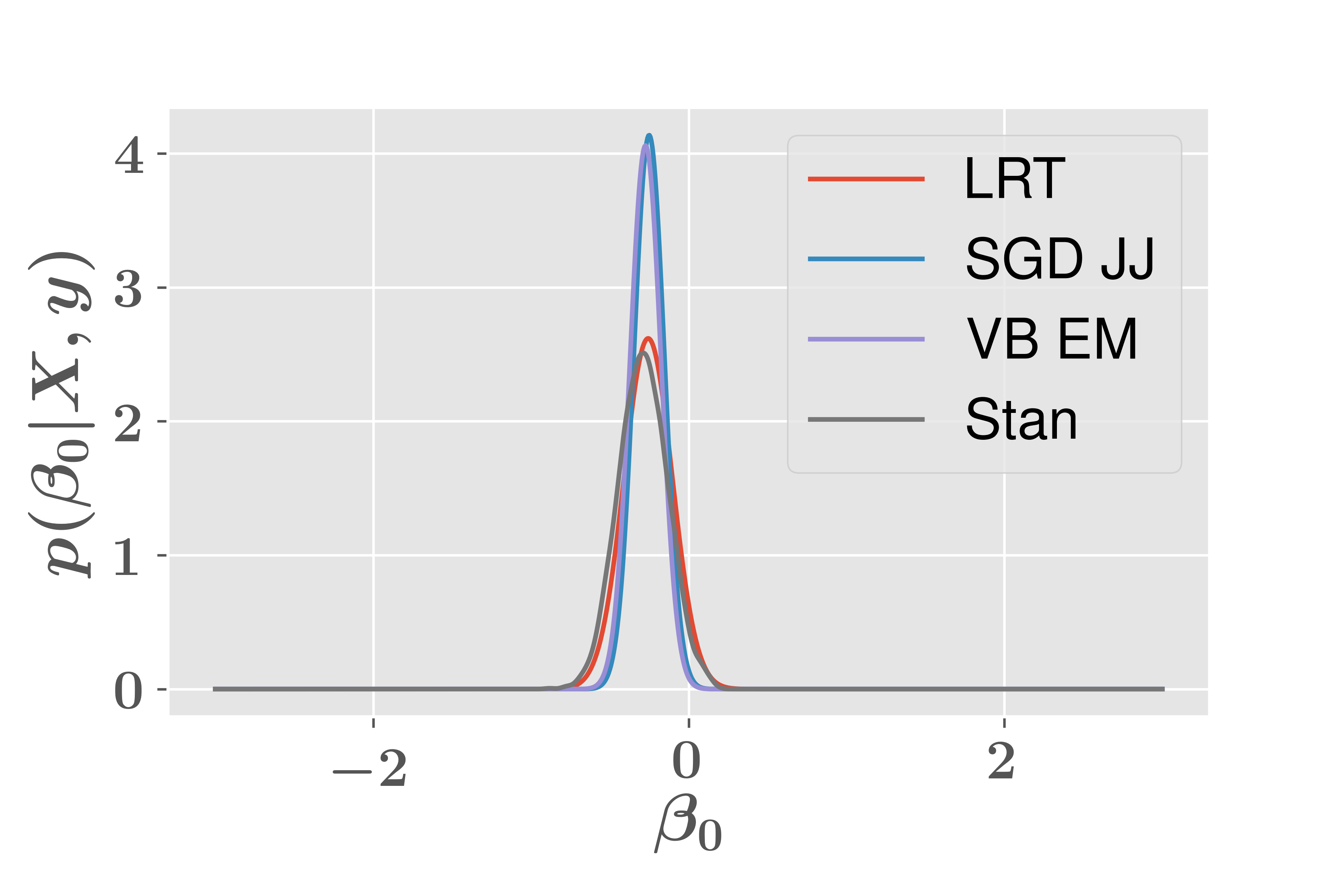}
  }
  \subfloat{
    \includegraphics[width=60mm]{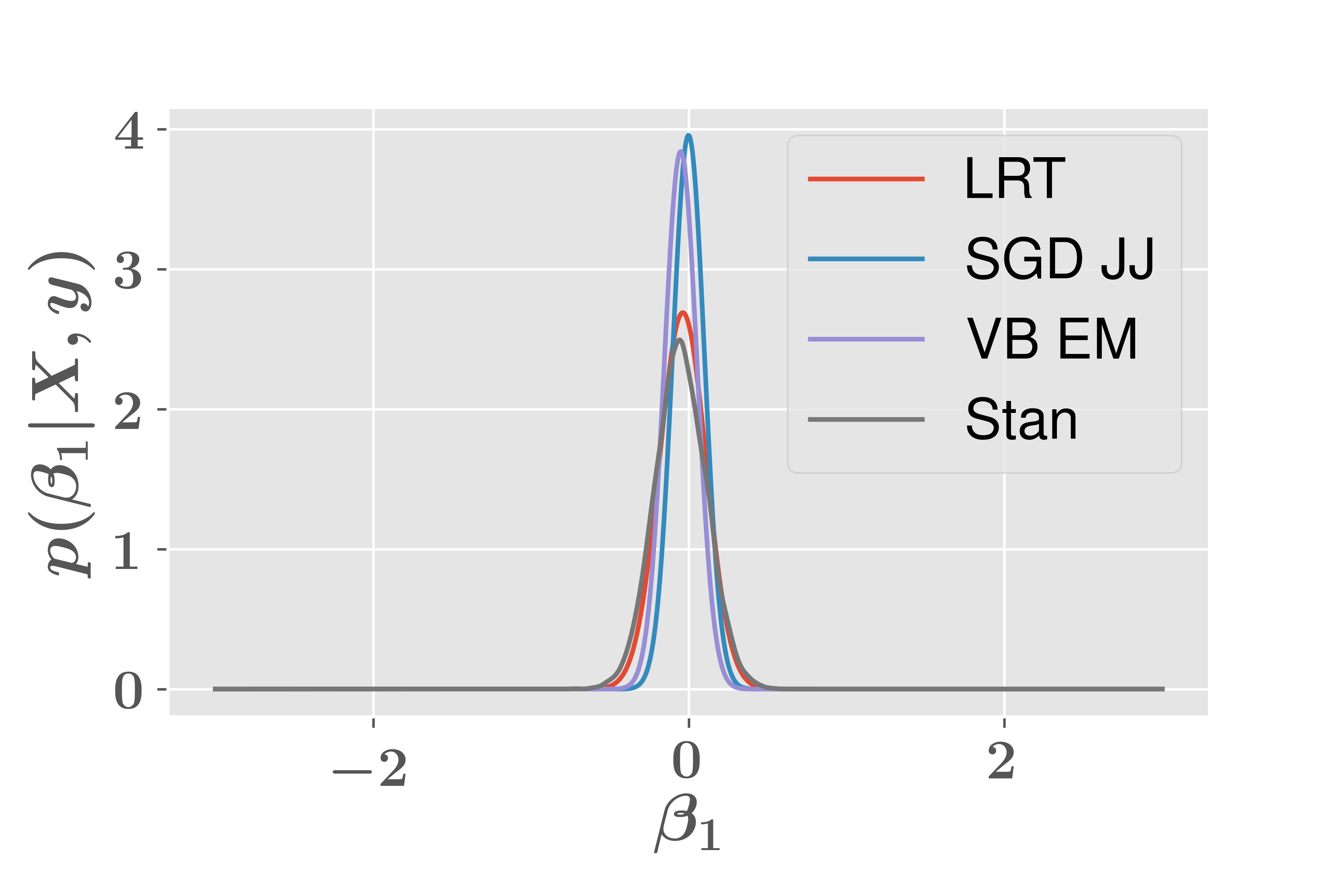}
  }
  \hspace{0mm}
  \subfloat{
    \includegraphics[width=60mm]{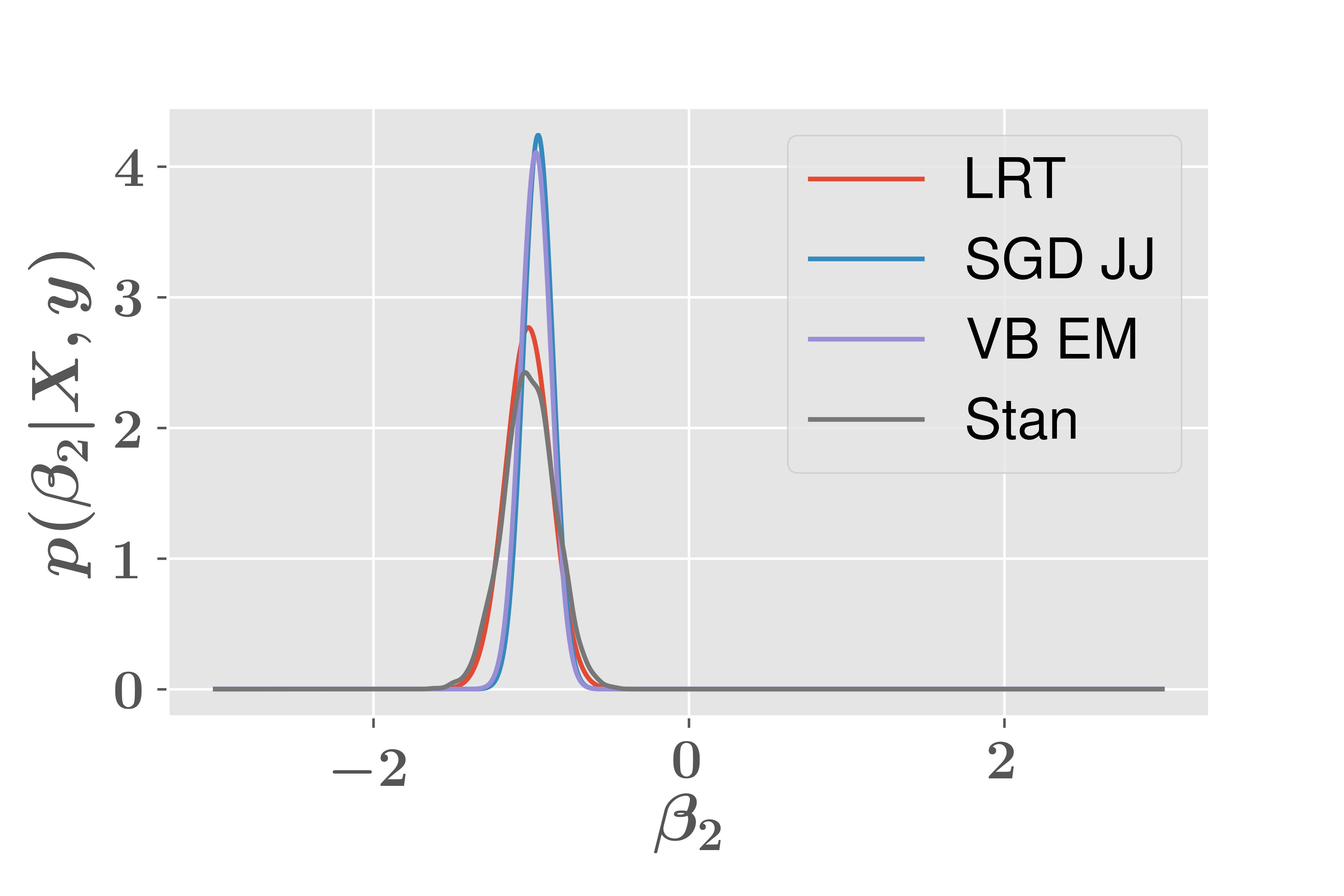}
  }
  \subfloat{
    \includegraphics[width=60mm]{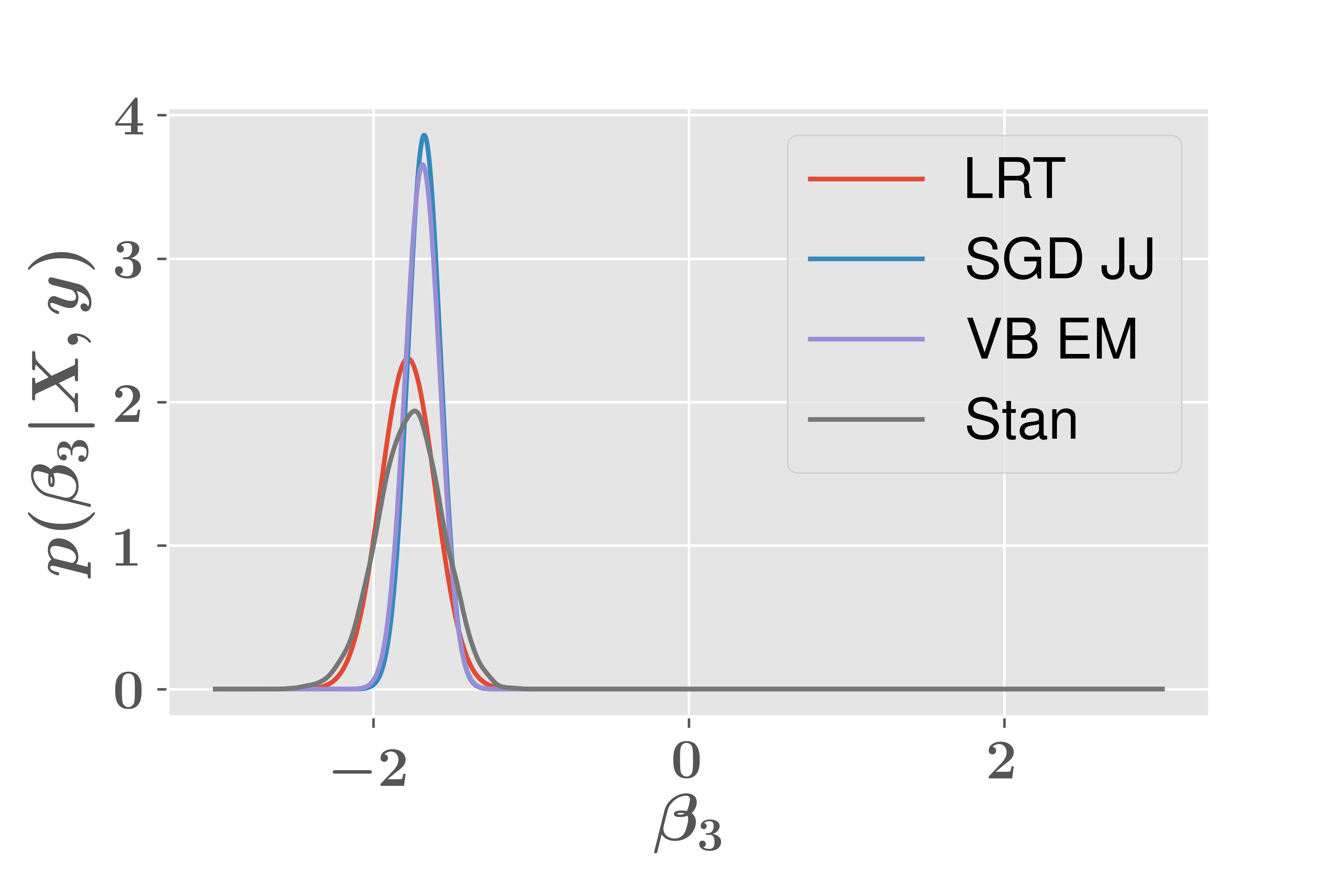}
  }
  \hspace{0mm}
  \subfloat{   
    \includegraphics[width=60mm]{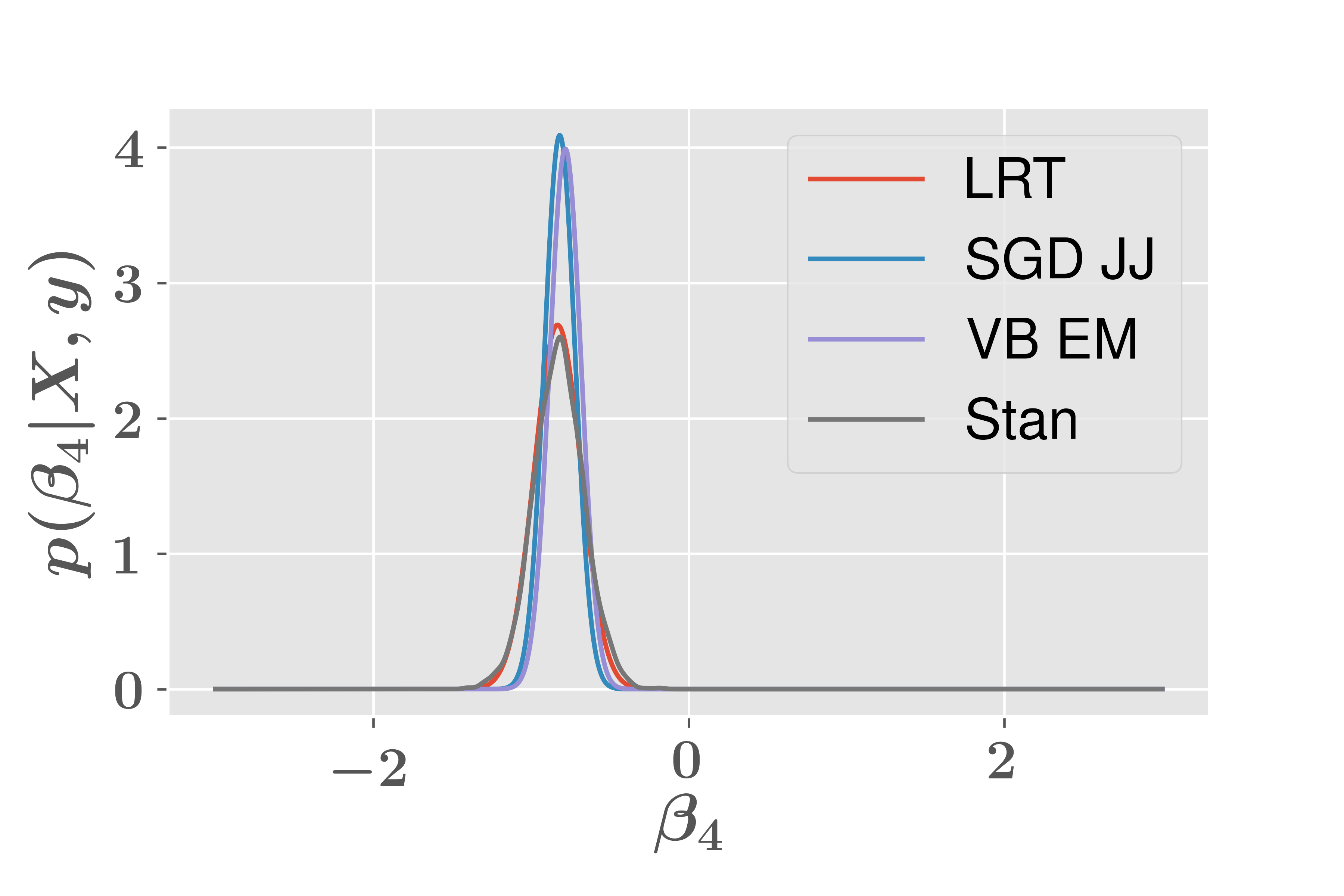}
  }
  \subfloat{
    \includegraphics[width=60mm]{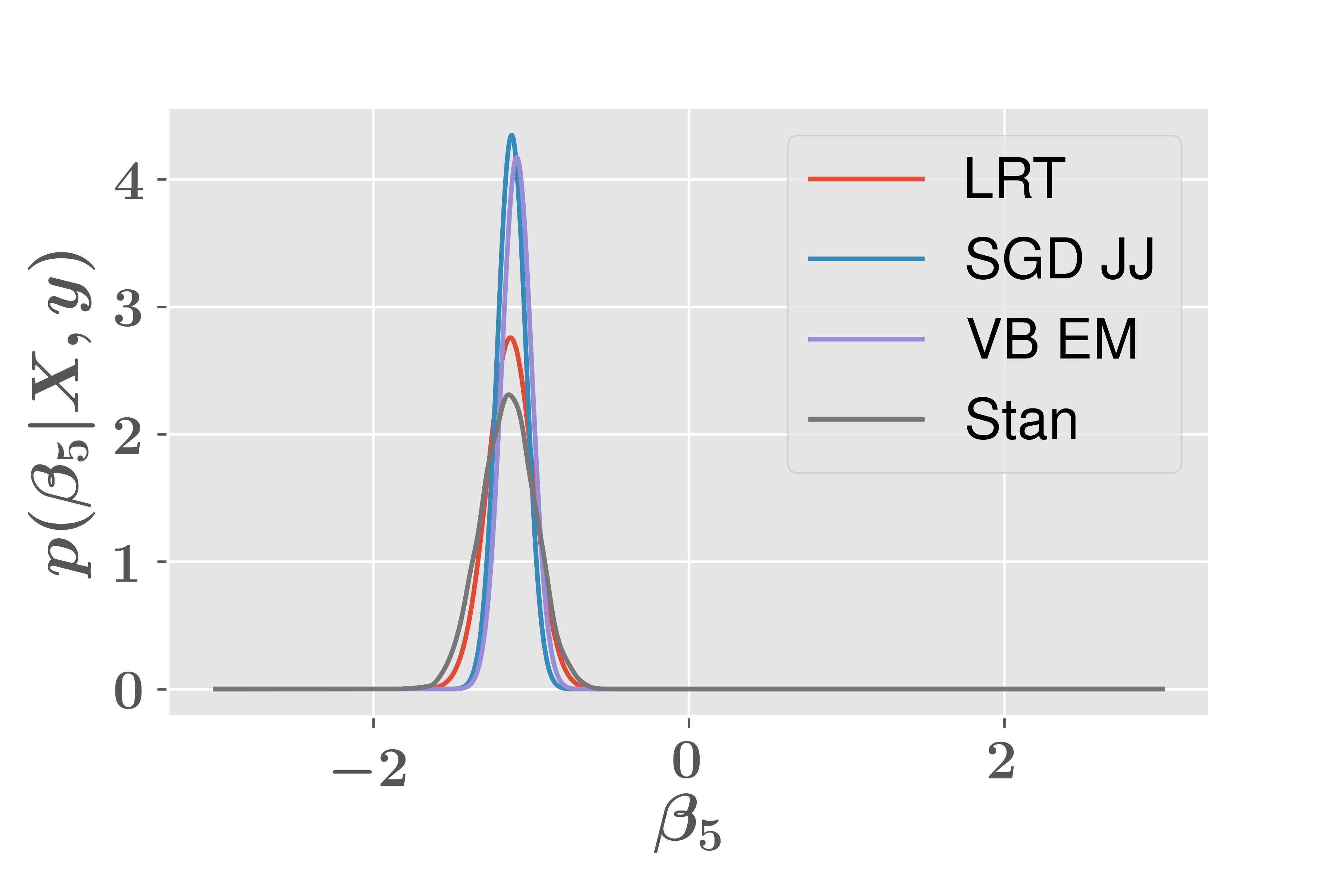}
  }  
  \caption{Posterior approximations for $\beta_0,..,\beta_5$.}
  \label{posterior_visualisation}
\end{figure}

In order to test the accuracy of our method, we simulated a logistic regression dataset of size 900 where $\bX$ has 50 features, 100 samples are held out for validation. We compute an approximate posterior using the Stan probabilistic programming language \cite{carpenter2017stan} which we take to be the gold standard, we also compute a posterior using the variational EM algorithm (the original use of the Jaakola and Jordan bound) (VB EM), the Local Re-parameterization Trick (LRT) and our proposed method Jaakola and Jordan SGD (JJ SGD). We take a kernel density estimate of the MCMC samples and plot the marginal posteriors for $\beta_0,...,\beta_5$ in Figure \ref{posterior_visualisation}.  It is apparent that all variational methods capture the mean well, but underestimate the posterior variance. The method that best captures the posterior variance is the local reparaemterization trick.  The stochastic gradient descent Jaakola and Jordan  and the variational Bayes EM algorithm - both of which use the Jaakola and Jordan bound underestimate the variance by a similar amount.  There is no obvious benefit for the use of the full covariance matrix used by VB EM as it has a similar level of fit to the true posterior as our proposed method SGD JJ.  SGD JJ is good at capturing the mean but it is worse than LRT at capturing the variance and has similar performance to VB EM.

\begin{figure}[H]
  \centering
    \includegraphics[width=70mm]{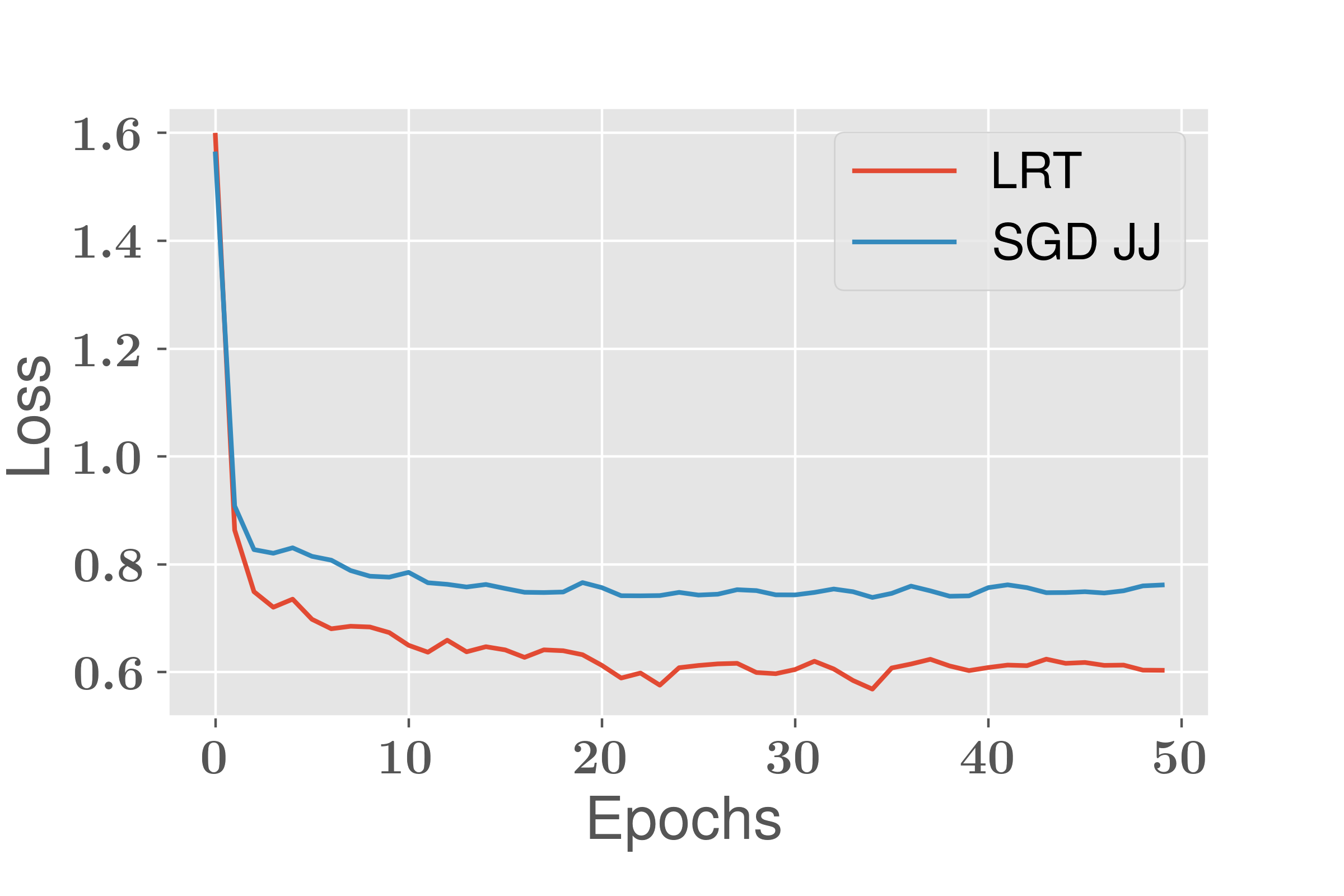}
  \caption{Loss curve of LRT and SGD JJ}
  \label{loss}
\end{figure}

In order to test the speed of the method we consider a harder problem and benchmarked only with LRT as the most scalable alternative.  We again simulate a logistic regression problem this time with 9000 records and 2000 features.  The cost per iteration of both methods are quite similar, each running on a CPU we get about 1 iterations per second.  The loss curve of the two methods are shown in Figure \ref{loss}, due to the Monte Carlo noise in the loss the LRT is slightly more difficult to optimize than SGD JJ and SGD JJ reaches its (higher) loss after typically fewer epochs.

Broadly we conclude that JJ SGD is less accurate than LRT, it iterates at the same speed and due to the fact it doesn't use Monte Carlo methods it has a less noisy loss, which in some situations allows faster convergence, although LRT may also be made less noisy by the use of Polyak Ruppert averaging.

\subsection{Bouchard Softmax Latent Variable Model Variational Autoencoder}

We evaluate the session based recommendation algorithm using data simulated from the RecoGym simulator \cite{rohde2018recogym}.  RecoGym is an environment for testing recommendation algorithms in an interactive environment applying reinforcement learning and bandit style evaluation to recommendation.  We use the simulator with 1000 products, we sample 200 user timeliness for training and 100 for testing.  We train our model both using the noisy softmax partition function approximation (sampling 200 products) and also without using the noisy softmax approximation i.e. (summing over all 1000 products without sampling).  It is notable that the speedups from approximating the partition function are limited by fixed costs required for processing the numerator of the softmax i.e. the ``positive examples'', for large numbers of products the ``negative sampling'' variate iterates approximately three times faster.   We use a latent factor size of 200, the variational auto-encoder is linear with the means unconstrained and the variances and $a$ parameters coming from a softplus transform.  The covariance matrix is constrained to be diagonal.  We compare the method with some simple recommendation baselines: popularity (Pop) a non-personalized recommendation strategy that recommends the most popular products to everybody, Item k-nearest neighbors (Itemknn) we estimate the empirical correlation matrix and then recommend the five items that is most correlated to the most recently viewed items.  Finally we also present results training the model using the classic Kingma and Welling algorithm that does not employ the Bouchard bound.  All results are shown in Table \ref{table2}.  The metrics presented are recall at 5 and truncated discounted cumulative gain at 5 (see \cite{liang2018variationalold} for a definition).  We see the re-parameterization trick performs the best closely followed by models utilizing the Bouchard bound using the full partition function and using the noisy approximation of the partition function respectively.

\begin{table}[H]
  \center
\begin{tabular}{llrr}
  \toprule
  Algorithm &   Recall@5 &  TDCG@5 \\
    \midrule
  Itemknn &  0.088 &   0.116 \\
  Pop &  0.090 &   0.090 \\
  Bouch/AE & 0.179 &   0.201\\
  Bouch/AE/NS & 0.165 &   0.191\\
  RT &  0.208 &   0.233  \\
    \bottomrule
\end{tabular}
\vspace{10pt}
\caption{Results for a model trained on 200 RecoGym user time-lines with 1000 products.  Test set size is 1000 user time-lines.}
\label{table2}
\end{table}  

\section{Conclusion}

In this paper, we have studied the use of analytical variational bounds in a modern deep learning setting, using auto-encoders or analytical EM steps to write the model as a sum so it can be trained using SGD. It is noteworthy that in this setting the variational auto-encoder doesn't necessarily have a classic ``auto-encoding'' interpretation, rather it is a dimensionality reduction technique that causes a variational bound with parameters growing with the dataset size to have a restricted dimension. The method applies both to latent variable models and to other methods that are not normally viewed in a latent variable setting such as logistic regression.

A significant advantages of the proposed method is primarily that a Bayesian approximation requires nothing more than SGD based optimization.  Both the Jaakola and Jordan and auto-encoding Bouchard bound were shown to be viable approximations for Bayesian logistic regression and the latent variable session model respectively. The circumstances where the proposed method performs well or badly with respect to alternatives, such as the local re-parameterization trick, remains a subject of further work. Clearly the re-parameterization trick and the local re-parameterization trick provide very strong baselines in terms of both accuracy and speed.

A further advantage of the use of the Bouchard bound is the ability to do a fast approximation of the partition function by an algorithm that resembles the ``negative sampling'' heuristic but is motivated in a fully probabilistic setting.  This method has promise in allowing fully probabilistic models to be applied to categorical variables with large number of classes.

\bibliographystyle{ACM-Reference-Format}
\bibliography{literature}

\end{document}